\def\year{2021}\relax
\documentclass[letterpaper]{article} 
\usepackage{aaai21}  
\usepackage{times}  
\usepackage{helvet} 
\usepackage{courier}  
\usepackage[hyphens]{url}  
\usepackage{graphicx} 
\usepackage{natbib}  
\usepackage{caption} 
\usepackage{booktabs} 
\usepackage[list=true]{subcaption}
\usepackage{threeparttable}
\usepackage{hyperref}

\title{Two Novel Performance Improvements for Evolving CNN Topologies}
\author{
    Yaron Strauch,
    Jo Grundy \\
}
\affiliations{
    University of Southampton\\
    y.strauch@soton.ac.uk, j.grundy@soton.ac.uk
}
\begin{document}

\maketitle

\begin{abstract}
Convolutional Neural Networks (CNNs) are the state-of-the-art algorithms for the processing of images. However the configuration and training of these networks is a complex task requiring deep domain knowledge, experience and much trial and error. 
Using genetic algorithms, competitive CNN topologies for image recognition can be produced for any specific purpose, however in previous work this has come at high computational cost. 
In this work two novel approaches are presented to the utilisation of these algorithms, effective in reducing complexity and training time by nearly 20\%. This is accomplished \textit{via} regularisation directly on training time, and the use of partial training to enable early ranking of individual architectures.
Both approaches are validated on the benchmark CIFAR10 data set, and maintain accuracy.
\end{abstract}

\section{Introduction}
\emph{Convolutional Neural Networks} (CNNs) consist of a set of layers of different types to sequentially pipe and transform data with. Early \emph{convolutional layers} detect edges or corners using \emph{filters}, \emph{pooling layers} reduce the resolution using an \emph{average} or \emph{max} function, allowing later layers to recombine smaller features into bigger ones. \cite{krizhevsky2009learning,krizhevsky2012imagenet}

The weights of the filters can be found through back propagation and gradient descent \cite{lecun1998gradient,lecun1989backpropagation}. However the overall architecture of the CNN must be stated, and many hyperparameters initialised, these are set by hand using trial and error 
\cite{simard2003best,ioffe2015batch}. This can introduce a bias towards benchmark data sets such as MNIST, which may have little relevance to real world data sets. Manual tuning of deep architectures is complicated, non-intuitive and time consuming.  

A promising approach to find topologies automatically is \emph{Genetic Algorithms} (GAs), heuristic 
algorithms that can optimise within abstract high-parameter spaces with complex or unknown interdependencies \cite{back1996evolutionary,davis1991handbook}. GAs implement a simplified model of 
natural selection. A population of individuals is initialised, where each individual is represented by a \emph{genome}. Their \emph{fitness} is measured by an \emph{objective function} that maps each genome to a fitness.
Fitness scores can be subjected to a \emph{penalty function} \cite{goldenberg1989genetic} that punishes undesired features or behaviour with selective pressure. 
Individuals of higher fitness are more likely to be selected for reproduction. \emph{Genetic operators} manipulate the genome stochastically: \emph{Mutation} is a small random change in the genome and \emph{cross-over} recombines the genome of two parents
\cite{Mooney1995}. Reproduction, genetic operations and fitness evaluation are repeated until an exit condition occurs.

CNNs can be configured through GAs by evolving network weights and topology together \cite{real2017large}. Children inherit the weights of their parents, mutations insert, alter or remove layers. CNNs are trained gradually during the course of evolution, allowing the objective fitness function to select individuals based on partially trained networks. The big drawback of this approach is that no cross-over is involved,  
reducing the potency of the algorithm by omitting an important operator of GAs. Introducing cross-over for weights is non-trivial and has not been achieved yet. 

Another approach is to evolve the topology alone, allowing cross-over. This method however requires retraining all weights for each topology configuration, before evaluation. To produce competitive results, algorithms can run for up to 35 GPU days \cite{sun2018automatically}.

In this work, two novel approaches are presented.
The first approach contrasts with previous work where complexity is penalised \cite{Kouvaris2017}, instead penalising evaluation wall time, allowing a more intuitive and direct effect.
Additionally, a method is presented to utilise both partial training and cross-over by gradually training models through an epoch function. Both approaches maintained accuracy on the \mbox{CIFAR10} image classification data set \cite{krizhevsky2009learning}.

\section{Method}
Initially a base experiment is set up, following Sun \cite{sun2018automatically}, with minor differences to aid computational efficiency. \footnote{The code is available at \url{https://github.com/YStrauch/evolving-cnn-topologies-for-image-recognition}} 

The GA evolves a generational population of 20 individuals, each represented by a genome describing the topology of a CNN. Each individual is evaluated by training and testing on the respective \mbox{CIFAR10} partition.

The initial genomes are generated by randomly concatenating skip or pooling layers with equal probability until a pooling layer would reduce the resolution to half a pixel, or the number of layers is greater than a max depth chosen randomly from $[10, 120]$. Skip layers contain two convolutional layers with quadratic filters of size 3, unit stride, and same padding. The number of filters of each convolutional layer is chosen randomly from the set $\{64, 128, 256\}$. Pooling layers use either a max or average function chosen randomly. A \emph{Multi-Layer Perceptron} (MLP) with no hidden layers followed by a softmax layer determines the output.
Individuals are trained for 60 epochs in a batch size of 50 using stochastic gradient descent with a momentum of .9 and an initial learning rate of .1, which decays by a factor of .9 after 1, 26 and 43 epochs.

CNNs are implemented using pytorch \cite{paszke2017automatic} and trained on 4 GTX1080 TI on the Iridis 5 HPC. Each GPU concurrently evaluates one individual at a time.
In the baseline implementation, the fitness of an individual is equal to its accuracy on the test fold. If an architecture was previously evaluated its accuracy is retrieved from a cache.

Individuals are selected for reproduction using tournament selection, \textit{i.e.} two distinct individuals are chosen randomly from the population, and the individual with higher fitness is selected. With a 90\% chance, this individual is crossed over with a second individual selected by another tournament selection. With a 20\% chance, the genome is mutated. Both genetic operators can be applied cumulatively, cross-over first. Selected individuals are added to a new generation of individuals until 20 individuals are found. If the best individual of the old generation is not in the new one, it replaces the weakest individual of the new generation (\emph{elitism}). The new generation replaces the old one. This process is repeated for 20 generations.

The one-point cross-over slices the feature detection layer stack open and re-combines them cross-wise. The Softmax-MLP classifier is not part of cross-over and is always appended once.
Mutation either inserts a skip layer with 70\% probability, or with each 10\% likelihood inserts a pool layer, removes a layer, or mutates a layer. 
Layers are inserted at a random position. Skip layers are mutated by re-randomising the number of filters; pool layers flip their kernel function between average and max.
If cross-over or mutation would produce half pixels due to too many pooling layers, the operator is aborted and restarted from the beginning.

\subsection{Regularisation}

The primary aim is to reduce the wall time taken by the algorithm. During fitness evaluation, the time required from the beginning of training to the end of testing is measured. For every hour spent, fitness is decreased linearly by $C$.

$C$ quantifies a trade-off between training time and accuracy, the bigger $C$, the higher the selective pressure towards faster individuals. This parameter was set at $C=0.05$ which allowed a reduction in training time without loss of predictive performance.

\subsection{Partial Training}

Evolution by natural selection does not need an absolute fitness score, but rather a relative measure to  rank each generation. The core idea is that earlier generations are trained for shorter durations, meaning less time is spent on earlier unfit topologies. In contrast to a previous approach \cite{real2017large}, weights are not heritable, so partial training cannot be done implicitly. Instead, an epoch function is introduced defining how long individuals are to be trained. To allow a fair comparison, partially trained individuals are stored so that they can be retrieved at a later time for further training.

The number of epochs each architecture is trained for is found using a linear function, dependent on the number of generations, ranging from a lower bound of 30 to an upper bound of 70, with rounding to the next integer. These bounds were chosen to make the algorithm spend less time on inferior and more time on better topologies. The learning rate decay points were scaled to come into effect after 1, 30 and 50 epochs.

\section{Results}

\begin{table}[tbh]
\centering
\begin{threeparttable}
\resizebox{.95\linewidth}{!}{
    \begin{tabular}{l|ccccc}
         \toprule
        Approach & Acc. & Gen. & Epochs & Batch & GPU \\
                 & \emph{(\%)}     &      &        & Size  & Days\\
        \midrule
        Evolving & 95.22 & 20  & 350  & 1 & 35 \\
        Topologies \tnote{a} & & & & &\\
        \midrule
        Regularised \tnote{b} & 89.06  & 20 & 60 & 50 & 14 \\
        \midrule
        Partial & 88.70  & 20 & 60 & 50 & 12 \\
        Training \tnote{b} & & & & &\\
        \midrule
        Base & 88.36  & 20 & 60 & 50 & 15 \\
        Experiment \tnote{b} & & & & &\\
        \midrule
        Evolving & 77.19 & 50 & 180 & 1 & 17 \\
        Weights \tnote{c} & & & & &\\
        \bottomrule
    \end{tabular}
}
\footnotesize
\begin{tablenotes}
\item[a] \cite{sun2018automatically}
\item[b] This work
\item[c] \cite{xie2017genetic}
\end{tablenotes}
\caption{Both performance improvements yielded better accuracies in less time than the base experiment.  
}
\label{tab:exp:comparison1}
\end{threeparttable}
\end{table}

\begin{figure}[!ht]
\begin{subfigure}{.25\linewidth}
  \centering
  \includegraphics[width=.95\linewidth]{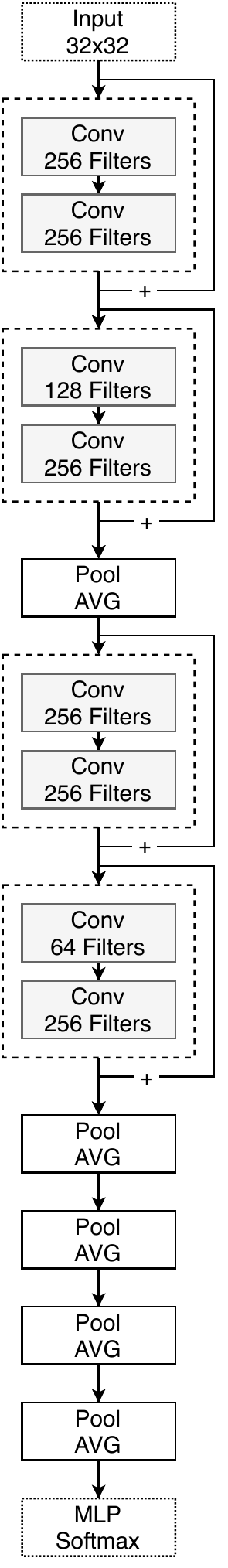}
  \caption{}
  \label{fig:architectures:base}
\end{subfigure}\hfill
\begin{subfigure}{.25\linewidth}
  \centering
  \includegraphics[width=.95\linewidth]{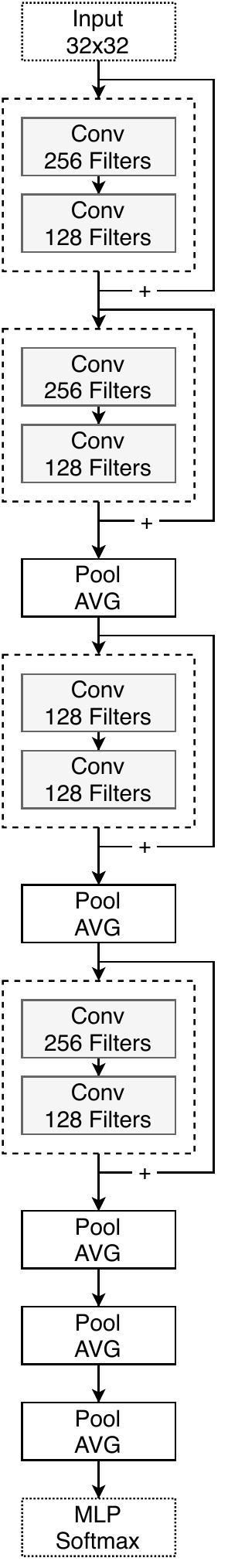}
  \caption{}
  \label{fig:architectures:reg:fitness}
\end{subfigure}\hfill
\begin{subfigure}{.25\linewidth}
  \centering
  \includegraphics[width=.95\linewidth]{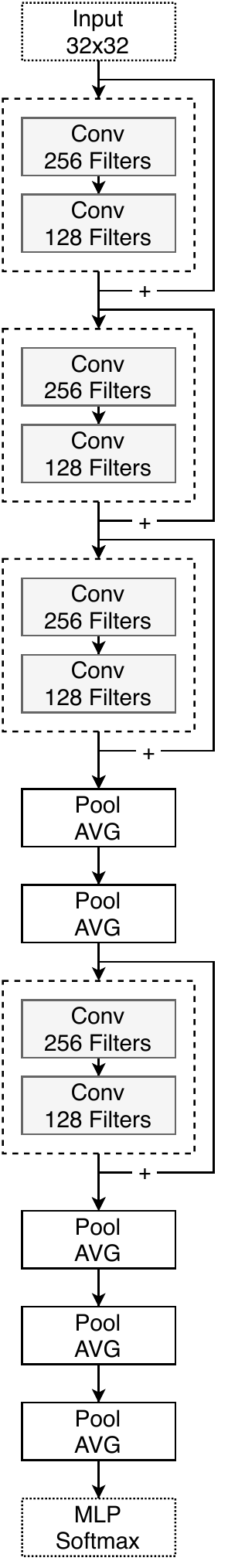}
  \caption{}
  \label{fig:architectures:reg:accuracy}
\end{subfigure}\hfill
\begin{subfigure}{.25\linewidth}
  \centering
  \includegraphics[width=.95\linewidth]{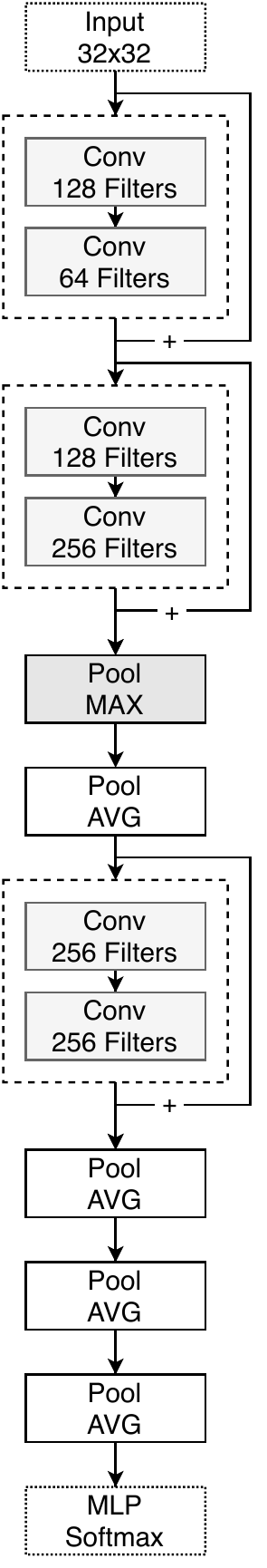}
  \caption{}
  \label{fig:architectures:epochfn}
\end{subfigure}\hfill
\caption{(\subref{fig:architectures:base}) The best of the base experiment. (\subref{fig:architectures:reg:fitness}) Regularised topology with highest fitness has fewer filters, and an extra pool layer. (\subref{fig:architectures:reg:accuracy}) With regularisation the best has more filters in one convolutional layer and has put two pool layers together. (\subref{fig:architectures:epochfn}) With partial training the best has fewer filters in the early layer and more later. Unlike the others, it includes a max filter.}
\label{fig:architectures}
\end{figure}

Table \ref{tab:exp:comparison1} compares two approaches from the literature with the base experiment and performance optimisations. The best architectures found by each approach are depicted in Figure \ref{fig:architectures}.
The best CNN found in the base experiment (Fig. \ref{fig:architectures:base}) has lower accuracy than that reported by Sun, however as theirs is trained for nearly 9 times as many epochs and over double the GPU days, this is unsurprising.
As methods are being compared, not time or resources available, it is this base experiment result that is used for accuracy and time comparison.

\begin{figure}[thb]
  \centering
  \includegraphics[width=.55\linewidth]{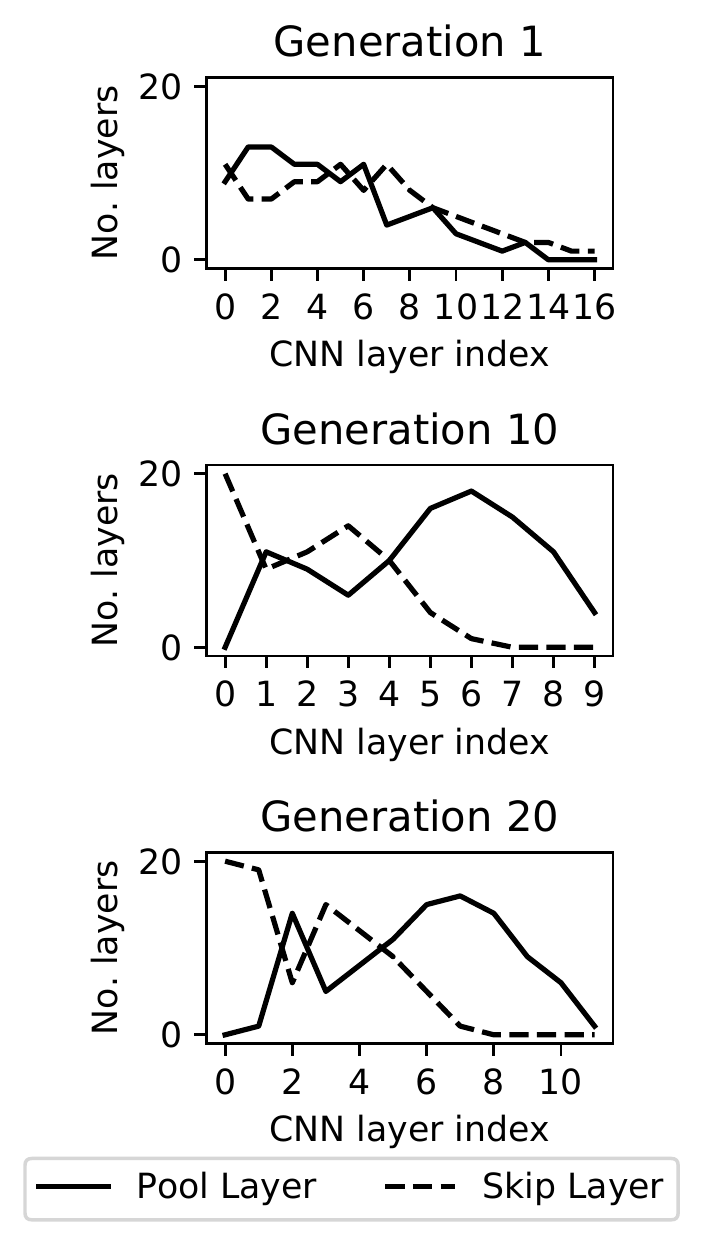}
  \caption{The population converges to have more skip layers in the first and more pooling layers in the second half of the genome. In generation 10 and 20, all individuals start with a skip layer. In generation 20, there is a pattern of alternating skip and pooling layers at depths 1-6.}
  \label{fig:exp1:layers}
\end{figure}

All the best topologies found have a stack of pooling layers in the end. Initially this was suspected to be due to the initialisation logic stopping when half pixels would occur, however analysis of the distribution of layers throughout evolution (Fig. \ref{fig:exp1:layers}) found an active drift towards this phenomenon. Remarkably, the population learned to alternate skip and pool layers, as shown in generation 20, Figure \ref{fig:exp1:layers}. The distribution curves meet three times between the first and fifth layer index. Alternating pool and skip layers is a pattern found in hand-crafted networks, \textit{e.g.} \cite{lecun1998gradient}, however the best topology found in this work arranges skip layers in pairs (see Fig. \ref{fig:architectures:base}). This is not a common design pattern used by humans.

\subsection{Regularisation}
\begin{figure}[thb]
    \begin{subfigure}[t]{.5\linewidth}
      \vspace{0pt} 
      \includegraphics[width=\linewidth]{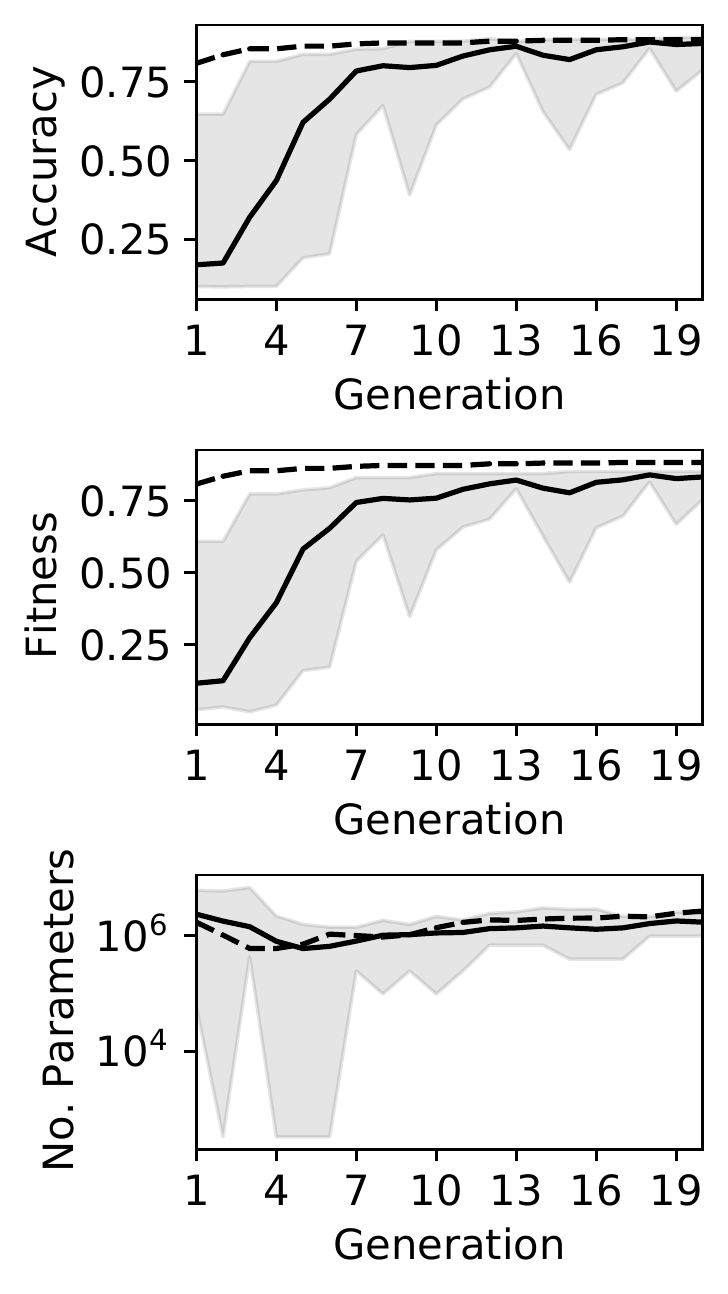}
      \caption{}
      \label{fig:exp5:stats}
    \end{subfigure}\hfill%
    \begin{subfigure}[t]{.5\linewidth}
		\vspace{0pt} 
    	\begin{subfigure}[t]{\linewidth}
      		\includegraphics[width=\linewidth]{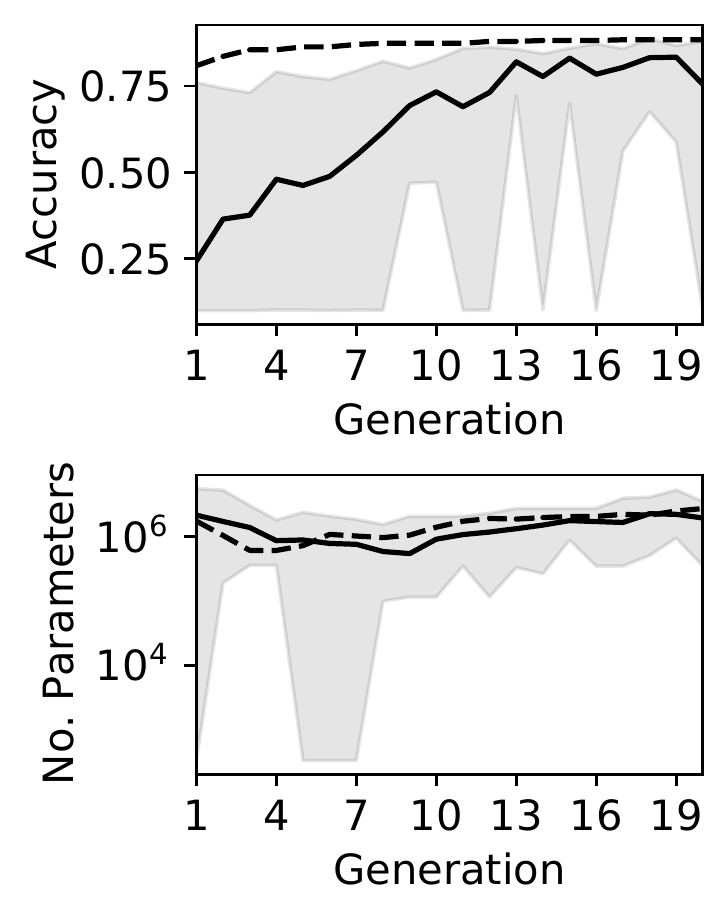}
      	\end{subfigure}\hfill%
      	\begin{subfigure}[b]{\linewidth}
	    	\vspace{15pt} 
      		\hfill 
      	 	\includegraphics[width=.6\linewidth]{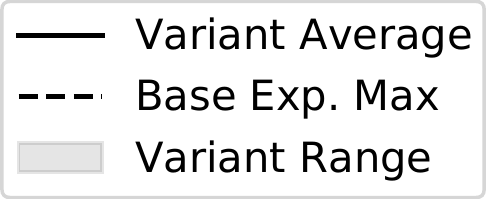}
	    	\vspace{15pt} 
      	\end{subfigure}
      \caption{}
      \label{fig:exp6:stats}
    \end{subfigure}
    \caption{(\subref{fig:exp5:stats}) The regularised networks converge to less complexity with comparable accuracy. Fitness is naturally lower due to the regularisation. 
    (\subref{fig:exp6:stats}) The partial trained accuracy converges more slowly in respect to generations, and catches up in generation 17. Reduced training will give lower fitness nearer the start.}
\end{figure}

The wall time of the regularised approach was 31 hours (8\%) shorter and yielded an accuracy that is slightly better than the base experiment. Convergence was compared (Fig. \ref{fig:exp5:stats}) and found to be similar overall. There is a constant, short delay on the X axis. As the first generation is more complex and less accurate, it seems likely that this is rooted in random chance within the initial genome generation, and is independent of regularisation.

Networks are 1-2 layers shallower on average, with comparable accuracy, fitness evaluations are naturally lower due to regularisation. The algorithm was therefore successful in discovering deep networks that performed as well as shallower networks. The spread in fitness evaluations does not collapse, and converges reasonably, indicating a good population diversity.

The regularised run favoured having fewer filters per skip layer. In contrast to the base experiment, the number of filters converged towards 128 instead of 256 filters. This indicates that having fewer filters increases training speed and maintains accuracy, at least for the few training epochs configured.

The individual with highest fitness (Fig. \ref{fig:architectures:reg:fitness}) and the one with highest accuracy (Fig. \ref{fig:architectures:reg:accuracy}) are very similar. The only differences are that the faster (fitter) individual uses 128 instead of 256 filters in one skip layer, and the order of this layer and its successor. These tiny changes allowed the individual to evaluate with 15 minutes difference and only sacrificed 0.0064 accuracy. Such small changes add up significantly in the course of the GA and allowed it to evaluate 8\% faster than the base experiment without sacrificing accuracy of the end result, the highest accuracy found is slightly higher.

\subsection{Partial Training}
\begin{figure}[thb]
  \centering
  \includegraphics[width=.6\linewidth]{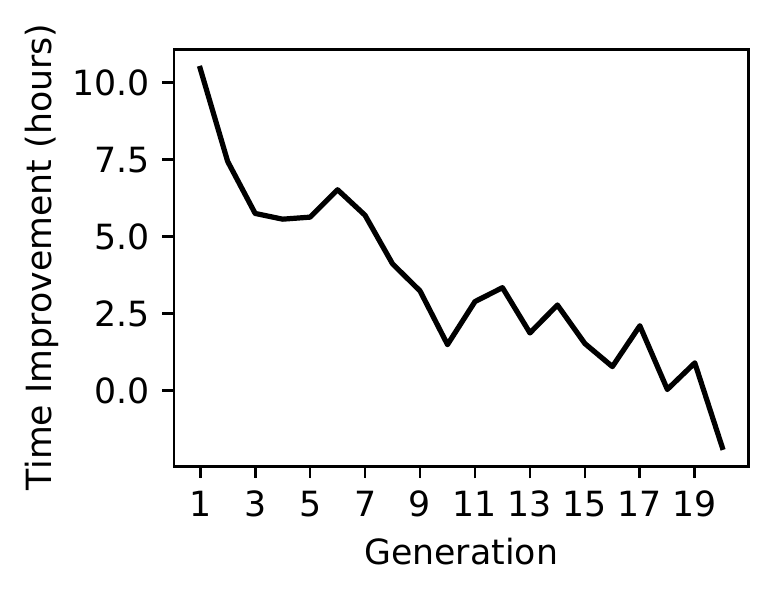}
  \caption{The speed-up of partial training is higher in earlier generations. Only the last generation took longer.}
  \label{fig:exp6:timediff}
\end{figure}

The GA ran 70 hours (19\%) faster than the base experiment. The best topology found (Fig. \ref{fig:architectures:epochfn}) has a slightly higher accuracy than the one found in the base experiment.
Figure \ref{fig:exp6:stats} shows a bigger spread in accuracy than the previous experiment, as the population is more diverse. The population converges more slowly in terms of generations, however this does not mean that convergence was slower in terms of wall time; on the contrary it actually converged much faster. As expected, the speed-up (Fig. \ref{fig:exp6:timediff}) is inversely proportional to the number of training epochs with some noise. The first two generations were evaluated 18 hours faster while maintaining the mean fitness of the base experiment (shown in Fig. \ref{fig:exp6:stats}), demonstrating the efficacy of this idea.
The same mean fitness as the base experiment was reached by generation 18, despite having saved more than 60 hours of computation (70 hours in total). Speed-up occurs even after generation 15 where the epoch function passes the baseline experiment of 60 epochs.

\section{Discussion and Conclusion}

Two methods to speed up convergence of genetic algorithms finding CNN topologies for image recognition are presented. 
Though results may vary depending on the hardware used, the advantage of using wall time is that selective pressure will drive the population to accommodate the performance of the underlying architecture. However, care should be taken not to mix different GPU types or individuals will be ranked unfairly. Also one has to keep in mind that individuals may be assigned a negative fitness score, which would complicate other genetic selection algorithms such as roulette selection.

The second approach changed training epochs over the course of evolution to allow removal of weak configurations early on without excessive training. The linear function presented uses a lower and upper bound to simplify migration from the flat value used in the base experiment. Changing this function to a more dynamic approach, such as reacting to previous fitness evaluations to train longer if needed could allow adaptation to a variety of datasets.

Both approaches are demonstrated to speed up architecture selection in this work. While they both introduce their own hyperparameters, the ability of these algorithms to tune numerous other parameters makes them very valuable, allowing the wider adoption of deep learning methods.

Interestingly, combining both approaches neither results in a combined speed-up nor a comparable accuracy. A speed-up of only 0.6\% compared to partial training, and top accuracy of 79.47\% (10\% worse than the base experiment) indicate that the two methods conflict with or even oppose another. This requires further investigation.

Further validation of the two approaches presented with their speed-ups, extensive repetition of the two experiments, and application to more data sets and use cases would be worthwhile. 
This work sets out two approaches new to this particular domain, and demonstrates their utility. 

\section{ Acknowledgments}
The authors acknowledge the use of the IRIDIS High Performance Computing Facility.

\bibliography{sources.bib}

\end{document}